\pdfoutput=1

\documentclass[11pt]{article}

\usepackage{authblk}
\usepackage{naacl2021}

\usepackage{times}
\usepackage{subfig}
\usepackage{latexsym}
\usepackage{multicol}
\usepackage{multirow}
\usepackage{booktabs}
\usepackage{graphicx}
\usepackage{booktabs}
\usepackage{mdframed}
\usepackage{comment}
\usepackage{amsmath,amsfonts,amssymb}

\usepackage[T1]{fontenc}

\usepackage[utf8]{inputenc}

\usepackage{microtype}

\usepackage{paralist}
\usepackage{outlines}
\usepackage{xspace}
\usepackage{caption}

\newcommand{\taskAbr}{ODQA\xspace}
\newcommand{\detectError}{error-detectability\xspace}

\newcommand{\eg}{\mbox{\it e.g.}}
\newcommand{\ie}{\mbox{\it i.e.}}

\title{Human Evaluation of Spoken vs. Visual Explanations \\ for Open-Domain QA}

\author[ \hspace{.1cm}1]{\textbf{Ana Valeria Gonz{\'a}lez}\thanks{\hspace{.2cm}Work done while at Facebook AI.}}
\author[2]{\textbf{Gagan Bansal}}
\author[3,4]{\textbf{Angela Fan}}
\author[3]{\textbf{Yashar Mehdad}}
\author[3]{\\\textbf{Robin Jia}}
\author[3]{\textbf{Srinivasan Iyer}}
\affil[ ]{$^1$University of Copenhagen, $^2$University of Washington, $^3$Facebook AI, $^4$LORIA}
\affil[ ]{\texttt{ana@di.ku.dk}}
\affil[ ]{\texttt{bansalg@cs.washington.edu}}
\affil[ ]{\texttt{\{angelafan, mehdad, robinjia, sviyer\}@fb.com}}

\begin{document}
\maketitle
\begin{abstract}

While research on explaining predictions of open-domain QA systems (\taskAbr) to users is gaining momentum, most works have failed to evaluate the extent to which explanations improve user trust. While few works evaluate explanations using user studies, they employ settings that may deviate from the end-user's usage in-the-wild: \taskAbr is most ubiquitous in {\em voice}-assistants, yet current research only evaluates  explanations using a {\em visual} display, and may erroneously extrapolate conclusions about the most performant explanations to other modalities. To alleviate these issues, we conduct user studies that measure whether explanations help users correctly decide when to accept or reject an \taskAbr system's answer. Unlike prior work, we control for explanation {\em modality}, \ie, whether they are communicated to users through a spoken or visual interface, and contrast effectiveness across modalities. Our results show that explanations derived from retrieved evidence passages can outperform strong baselines (calibrated confidence) across modalities but the best explanation strategy in fact changes with the modality. We show common failure cases of current explanations, emphasize end-to-end evaluation of explanations, and caution against evaluating them in proxy modalities that are different from deployment.

\end{abstract}

\section{Introduction}

\begin{figure}
\centering
    \includegraphics[width=.35\textwidth]{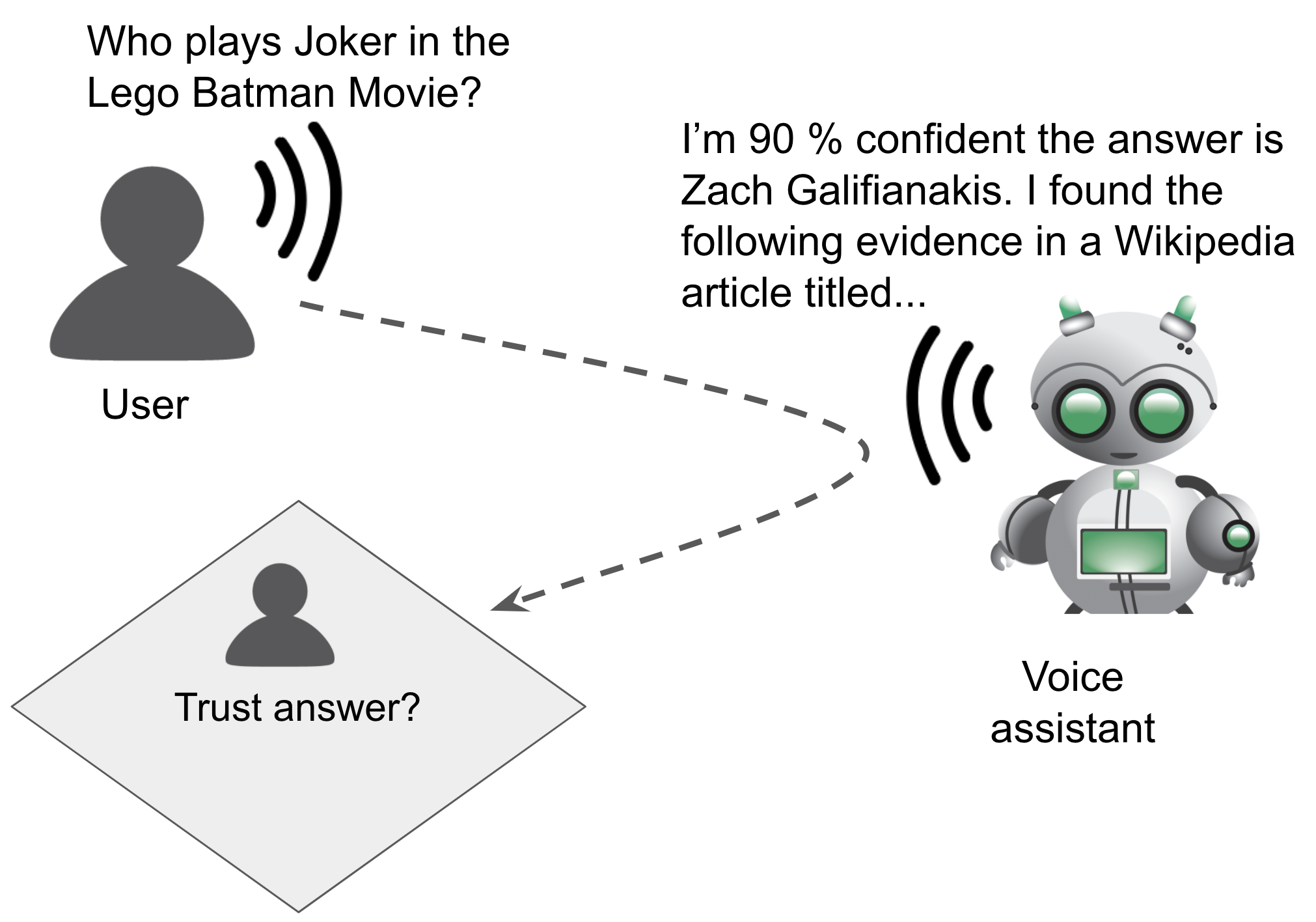}

    \caption{Using end-to-end user studies, we evaluate whether explanation strategies of open-domain question answering assistants help users decide when to trust (or reject) predicted answers.}
\end{figure}

Despite copious interest in developing explainable AI and NLP, there is increasing skepticism about whether explanations of system predictions provide value to users in many important downstream applications. For instance, for classifying sentiment and answering LSAT questions, \newcite{bansal2020does} observed that textual explanations were no more helpful to users for decision-making than simply presenting model confidence. Similarly, \newcite{chu2020visual} observed that visual explanations fail to significantly improve human accuracy or trust. Such negative results present a cautionary tale for explainable NLP and emphasize the need to evaluate explanations using well-designed user-studies.

In this work, we explore this problem for {\em \textbf{O}pen-\textbf{D}omain \textbf{Q}uestion \textbf{A}nswering} models, which are increasingly deployed in voice-assistant and Web search to answer users' questions.
In \taskAbr, users ask factoid questions (\eg, \textit{``Who plays the Joker in the Lego Batman movie?''}) and the system answers from a large corpus of documents (\eg, Wikipedia).
Even though the accuracy of such systems is rapidly improving, deployed models are imperfect and can make mistakes, pointing towards a need to provide users with mechanisms (\eg, displaying uncertainty) that aid in {\em appropriate reliance} on these systems. 
This motivates one of our key research questions, {\em Does explaining the system's reasoning, help users better assess when to (dis)trust their predictions?}  We henceforth refer to a user's ability to distinguish correct and incorrect answers as {\em \detectError}.

Recent work conducted user studies and concluded that explanations improve \detectError for \taskAbr~\cite{lamm2020qed,feng-iui19}. \newcite{lamm2020qed} showed that providing an evidence sentence as well as coreference and entailment information, improves \detectError marginally. However, this study lacked a strong baseline that uses calibrated confidence and which on other domains has been shown to be effective. While \newcite{feng-iui19} did present model confidence along with highlighting of important attributes, and influential training examples, their evaluation setup is only suitable for visual displays.

In practice, \taskAbr systems are ubiquitous in voice assistants, with recent surveys indicating that voice is an increasingly popular method for asking questions on a smartphone.\footnote{\url{https://www.perficient.com/insights/research-hub/voice-usage-trends}}Spoken language interfaces can also improve accessibility of technology for people with visual impairments and reading disabilities. Explanations in the spoken modality pose unique challenges--- information will impose different cognitive demands depending on whether it is shown visually or by voice \cite{sweller2011cognitive,leahy2016cognitive}, which could make long explanations less effective in the spoken modality.

We focus on two central questions: 
\textbf{(1)} Do explanations help users discriminate between correct and incorrect model predictions across both spoken and visual modalities? and \textbf{(2)} Do the preferred explanation strategies depend on modality? 
We present the first study comparing both visual and spoken explanations for \taskAbr on \detectError, involving over 500 participants, and evaluate three types of natural language explanations: a retrieved evidence paragraph, a retrieved evidence sentence, and a human-written sentence that abstractively summarizes the evidence.

Our findings indicate that while natural language explanations help users in both modalities, they can mislead users into accepting incorrect predictions. Additionally, users prefer different explanation types in different modalities.
In the spoken modality, all explanation types outperform the strong baseline of presenting calibrated confidence scores, demonstrating that they are helpful to users.
Among the three explanation types, extractive sentence explanations are the most useful.
On the other hand, extractive explanations and human-written abstractive explanations frequently mislead users into trusting incorrect answers.
In contrast, longer extractive explanations are more effective than sentence-length explanations in the visual modality, which demonstrates the importance of tailoring explanations to the user interface.

\section{Related Work}
\label{sec:related}

\paragraph{Natural Language Explanations}
Recent work has introduced neural models that are trained to perform a task and output a natural language (NL) explanation. \citet{camburu2018snli} and \citet{rajani2019explain}, both introduce methods for training self-explaining models using free-form NL explanations collected from crowdsourced workers for natural language inference and common sense reasoning. \citet{atanasova2020generating} introduce a method for generating explanations for fact verification using human veracity justifications. 
\citet{lei2016rationalizing} introduced an approach for extracting \textit{rationales} by selecting phrases from the input text which are sufficient to provide an output. Rationales have since been introduced for various NLP tasks \cite{DeYoung:ea:19, chen2018learning, yang2018hotpotqa}. 

\citet{lamm2020qed} introduce QED explanations in \taskAbr consisting of the sentence containing the answer, coreference and entailment information. However, unlike free-form explanations or rationales, these explanations are too complex to adapt to the spoken modality.
In question answering, many current models provide an answer and a rationale (or \textit{extractive} evidence). We evaluate extractive evidences from a state-of-the-art \taskAbr model, along with human-written summaries. 

\paragraph{Evaluating Explanations}

The quality of NL explanations has previously been evaluated using automatic metrics that measure the agreement of explanations with human annotations  \cite{DeYoung:ea:19, paranjape2020information,swanson2020rationalizing,camburu2018snli,rajani2019explain}. It is not clear how these metrics reflect the usefulness of explanations in practice. As the goal of explainability is to make model decisions more predictable to \textit{human} end users, a more useful way of evaluating explanations is through human evaluation.

Some human evaluations have used proxy tasks to evaluate explanations \cite{hase2020evaluating,nguyen2018comparing}, however, \citet{buccinca2020proxy} showed that both subjective measures and proxy tasks tend to be misleading and do not reflect results in actual decision making tasks.

Within question answering, \citet{feng2019can} evaluate how expert and novice trivia players engage with explanations. \citet{lamm2020qed} evaluate how QED explanations help raters determine whether a model decision is correct or incorrect, and find marginal improvements on rater accuracy. Unlike these works, we simplify the presentation setup so that we can adapt explanations across different modalities. 
\citet{bansal2020does} observed that for sentiment analysis and answering LSAT questions, state-of-the art explanation methods are not better than revealing model confidence scores and they increase the likelihood of users accepting wrong model predictions. We compare confidence to various explanation strategies for \taskAbr, but unlike previous work, we use \textit{calibrated} model confidence.

\paragraph{Open-domain QA}
\taskAbr consists of answering questions from a corpus of unstructured documents\footnote{\url{https://trec.nist.gov/data/qamain.html}}. Currently, \taskAbr models consist of two components: (1) a document \textit{retriever} which finds the most relevant documents from a large collection and (2)  a machine comprehension model  or \textit{reader} component, which selects the answer within the chosen documents ~\cite{chen2017reading,das2018multi,lee2019latent,karpukhin2020dense}. Recent work focuses on identifying answers in Wikipedia~\cite{karpukhin2020dense} as well as the web~\cite{joshi2017triviaqa}, encompassing both short extractive answers~\cite{rajpurkar2016squad} and long explanatory answers~\cite{fan2019eli5}.

\section{Visual vs. Spoken Modalities}
\label{sec:modalities}
When presenting NL explanations to users, we must keep in mind that users typically process information differently across the spoken and visual modalities. In this section we discuss work in learning and psychology research,  which point to the differences motivating our evaluation.

\begin{enumerate}
    \item \textbf{Real-time processing}:  \citet{flowerdew1994academic} observe that one of the main differences in how people process spoken versus written information is linearity. When listening, as opposed to reading, information progresses without you. Readers, on the other hand, are able to go back and dwell on specific points in the text, skip over and jump back and forth \cite{buck1991testing,lund1991comparison}. Although in some scenarios it is possible to get spoken information repeated, it may not be as effective as re-reading (see below).
    
    \item \textbf{Recall of information}: People tend to recall less after listening versus reading \cite{osada2004listening}. \citet{lund1991comparison} found that for some listeners, listening to information again was not as effective as re-reading.  While advanced listeners benefited from listening multiple times, this was a controlled learning scenario simulating students learning classroom material; we would expect users in an \taskAbr setting to be slightly more passive. 
    
    \item \textbf{Effect on concentration}: The heavier cognitive load imposed by listening to information can make people lose concentration more easily. \citet{thompson1996can} found that optimal length for listening materials was around 30 seconds to 2 minutes. Beyond that, listeners would lose full concentration. When people interact with voice assistants they may be on the go, or may be surrounded by additional distractions not present in a learning environment. This in turn may make the optimal length of material (explanations, in our case) much shorter.
\end{enumerate}
 We argue that these differences in processing of spoken and written information can have tremendous consequences in the effectiveness of natural language explanations in \taskAbr. Our experimental setup is the first to address these differences.

\section{Experimental Setup}

We design our user study to evaluate explanation effectiveness for \taskAbr by varying two factors: {\em type} of explanation and {\em modality} of communication. We combine variations of each factor to obtain explanation conditions (Section~\ref{sec:types}) and obtain them using a state-of-the-art \taskAbr model~(Section~\ref{sec:questions}). We then deploy these conditions as HITs on Amazon Mechanical Turk (MTurk) to validate five hypothesis, each stating relative effectiveness of conditions at improving \detectError~(Section~\ref{sec:hyp}).
Since MTurk studies can be prone to noise, to ensure quality-control, we make and justify various design choices (Section~\ref{sec:ui}).

\subsection{Explanation Types and Conditions}
\label{sec:types}
\taskAbr models can justify their predictions by pointing to {\em evidence} text containing the predicted answer \cite{das2018multi,lee2019latent,karpukhin2020dense}. We experiment with two types of {\em extractive} explanations: 

\begin{itemize}
    \item {\sc extractive-sent}: Extracts and communicates a sentence containing the predicted answer as evidence. 
    
    \item {\sc extractive-long}: Extracts and communicates a longer, multi-sentence paragraph containing the answer as evidence.
    
\end{itemize}
    
While extractive explanations are simpler to generate, we also evaluate a third explanation type that has potential to  more succinctly communicate evidence spread across documents \cite{liu2019towards}.
\begin{itemize}
    \item {\sc abstractive}: Generates and communicates new text to justify the predicted answer.

\end{itemize}

\paragraph{Final explanation conditions}
For the \textit{voice modality}, we test five conditions, two baselines and three explanation types:  (1) {\sc baseline}: present only the top answer, (2) {\sc confidence}, a second, stronger baseline that presents the top answer along with the model's uncertainty in prediction, (3) {\sc abstractive}, (4) {\sc extractive-long}, and (5) {\sc extractive-sent}. 

In the \textit{visual modality}, we have 2 conditions corresponding to the {\sc extractive-long} and {\sc extractive-sent} explanation types. Here, we were primarily interested in contrasting these with the voice modality. Examples of our explanations can be found in Appendix \ref{sec:appendix1}.

\subsection{Hypotheses}
\label{sec:hyp}

We investigated five (pre-registered) hypothesis about the relative performance of various explanation conditions at improving \detectError, motivated by pilot studies and authors' intuitions. 

\begin{itemize}
    \item \textbf{H1}: Presenting model confidence will improve performance over the baseline.

    \item \textbf{H2}: \textit{Spoken} {\sc extractive-sent } explanations will perform better than {\sc confidence} --- the explanation would provide additional context to help validate predicted answers.

    \item \textbf{H3}: \textit{Spoken} {\sc extractive-sent } will perform better than \textit{Spoken} {\sc extractive-long}. Since the spoken modality may impose higher cognitive limitations on people~(Section~\ref{sec:modalities}), users may find concise explanations more useful despite them providing less context.

    \item \textbf{H4}: {\sc abstractive} will help users discriminate between correct and incorrect more than {\sc confidence} alone.

    Abstractive summaries may contain more relevant information than extractive explanations of the same length, which may help users make better decisions. 

    \item \textbf{H5}: \textit{Visual} {\sc extractive-long } will perform better than \textit{spoken} {\sc extractive-long }.
\end{itemize}

\begin{figure*}[h]
    \centering
    \includegraphics[width=\linewidth]{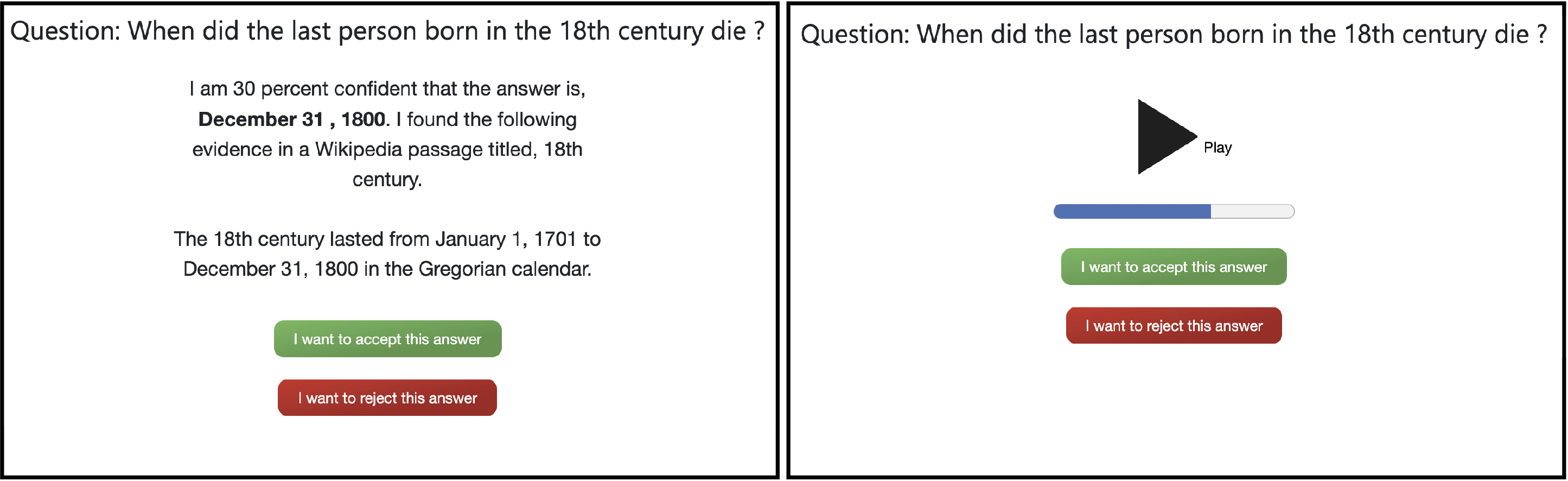}
    \caption{\textbf UI for visual (left) and spoken modalities (right) for {\sc extractive-sent} explanation type. Users either read or hear an explanation and decide whether to trust or discard the QA system's prediction.}
    \label{fig:ui-spoken}
\end{figure*}
    
\subsection{Implementation Details for Conditions}
\label{sec:questions}

\paragraph{Dataset}
For training our model and obtaining test questions for our user study, we used questions, answers, and documents
from the Natural Questions (NQ) corpus \cite{kwiatkowski2019natural}. NQ is composed of anonymized queries posed by real users on the Google search engine, and the answers are human-annotated spans in Wikipedia articles. The \textit{naturally occurring} aspect of this data makes it a more realistic task for evaluating explanations. To simplify the study, we restrict our attention to the subset of questions with short answers (< 6 tokens) following \newcite{lee-etal-2019-latent}. This subset contains 80k training examples, 8,757 examples for development, and 3,610 examples for testing.

\paragraph{Model}
We train the current (extractive) state-of-the-art model on NQ: Dense Passage Retrieval and Reader (DPR)~\cite{karpukhin2020dense}.
Similar to \citet{karpukhin2020dense}, we split documents (entire Wikipedia articles), into shorter passages of equal lengths (100 tokens). To answer an input question, DPR uses two separate {\em dense} encoders $E_Q(\cdot)$ and $E_P(\cdot)$ to encode the question and all passages in the corpus into vectors. It then retrieves $k$ most similar passages, where {\em passage similarity} to a question is defined using a dot product:  $sim(q,p) = E_Q(q)^{\intercal}E_P(p) $.

Given the top $k$ passages, a neural reader (Section~\ref{sec:related}) assigns a passage selection score to each passage, and a {\em span score} to every answer span. The original model uses the best span from the passage with the highest passage selection score as the final answer. However, we re-score each answer using the product of the passage and span score and use the highest-scored answer as the prediction. Our initial analysis showed that this re-scoring improved exact match scores of predicted answers.

\paragraph{Generating explanations}
To create our extractive explanations, we use the passage associated with DPR's answer--- {\sc extractive-sent} is defined as the single sentence in the passage containing the answer and {\sc extractive-long} is defined as the entire passage.
Since DPR does not generate abstractive explanations, we simulate {\sc abstractive} by manually creating a single sentence that captures the main information of {\sc extractive-sent} and adds additional relevant information from {\sc extractive-long}, whilst remaining the same length as {\sc extractive-sent}.

In order to improve transparency, in addition to presenting the evidence text in each explanation condition, we also inform users that the source of the text is Wikipedia and provide them with the \textit{title} of the article containing the passage together with the model's (calibrated) uncertainty in its prediction. Figure~\ref{fig:ui-spoken} shows an example of the final {\sc extractive-sent} explanation condition.

To convert text to speech, we use an internally-available TTS tool. 
For the questions we used in our study, when spoken, our final {\sc abstractive} and {\sc extractive-sent} conditions were on average 15 seconds long, {\sc extractive-long} was between 30-40 seconds.

\paragraph{Confidence calibration}
Confidence scores generated by neural networks (\eg, by normalizing softmax scores) often suffer from poor calibration\citet{guo2017calibration}.
To alleviate this issue and to follow best practices~\cite{amershi-chi19} for creating strong baselines, we calibrate our \taskAbr model's confidence using {\em temperature scaling}~\cite{guo2017calibration}, which is a  {\em post hoc} calibration algorithm suitable for multi-class problems. We calibrate the top 10 outputs of the model. We defer details on the improvement in calibration obtained through temperature scaling, and its implementation, to Appendix \ref{sec:temp-scale}.

\subsection{User study \& Interface} 
\label{sec:ui}
We conduct our experiments using Amazon Mechanical Turk. Our task presents each worker with 40 questions one-by-one, while showing them the model's answer (along with other condition-dependant information, such as confidence or explanation) and asks them to either {\em accept} the model's prediction if they think it is correct or {\em reject} it otherwise. Figure~\ref{fig:ui-spoken} shows an example.
For each of the 7 conditions we hire 75 workers.

Additional details about our setting and the instructions can be found in Appendix \ref{sec:appendix2}.

\paragraph{Question selection}

We deliberately sample a set of questions on which the model's aggregate (exact-match) accuracy is 50\%; thus any improvements in \detectError, beyond random-performance, must be a result of users making optimal assessment about the model's correctness. To improve generalization of results, we average results over three such mutually exclusive sets of 40 questions. Before sampling the questions we also removed questions that were ambiguous or questions where the model was indeed correct but the explanations failed to justify the answer. For brevity, we defer these details and justifications of these details to the Appendix~\ref{sec:preprocess}.

\paragraph{Incentive scheme} To encourage MTurk workers to engage and pay attention to the task,
we used a bonus-based strategy --- When users accept a correct answer, we give them a 15 cent bonus but when they accept an incorrect answer they lose the same amount\footnote{If participants ended up with a negative bonus, no deductions were made from their base pay, instead their bonus was simply zero}. This choice aims to simulates real-world cost and utility from interacting successfully (or unsuccessfully) with AI assistants~\cite{bansal2019beyond}. Table~\ref{tab:payoff} shows the final pay-off matrix that we used.

\begin{table}[h]
\begin{center}
\begin{tabular}{l|cc}
\toprule
{\sc Prediction/Decision}&{\sc Accept}& {\sc Reject}\\

\midrule 
{\sc Correct} & +\$0.15& \$0\\
{\sc Incorrect} & -\$0.15 & \$0\\

\bottomrule
\end{tabular}
\end{center}
\caption{\label{tab:payoff} MTurk worker's bonus as a function of the correctness of \taskAbr model's prediction and the user's decision to accept or reject the predicted answer.}
\end{table}

\paragraph{Post-task survey} After the main task, we asked participants to (1) rate the length of responses, (2) rate their helpfulness and (3) give us general feedback on what worked and how explanations could be made better. For the \textit{spoken modality}, we also asked participants to rate the clarity of the voice to understand if confusion originated from text-to-speech challenges. The survey as presented to the participants can be found in Appendix \ref{sec:survey}.

\paragraph{Quantitative measures of \detectError}

We quantify user performance at \detectError using the following five metrics:

\begin{itemize}
    \item \textbf{Accuracy}: Percentage of times a user accepts correct and rejects incorrect answers. A high accuracy indicates high \detectError.
    
    \item \textbf{Cumulative reward}: The total dollar reward in bonuses earned by a worker based on the payoff in Table~\ref{tab:payoff}. Note that, unlike accuracy, the payoff matrix is not symmetric wrt. user decision and correctness of predictions.
    
    \item \textbf{\% Accepts | correct}: Percentage of times the user accepts {\em correct} answers. If a setting yields a \textit{high} true positive rate, this would indicate this setting helps users better recognize correct model responses. 
    
    \item \textbf{\% Accepts | incorrect}: Percentage of times the user accepts {\em incorrect} answers. For this metric, \textit{lower} is better. If a setting yields a high number, this would indicate that this setting misleads users more often. 

    \item \textbf{Time}: The median time taken by a user to reach a decision after a question is shown. 
\end{itemize} 

When computing these metrics, we remove the first 4 questions for each worker to account for workers getting used to the interface. We pre-registered this procedure prior to running our final studies and, as as result, the maximum cumulative reward for this setup is \$ 2.70.

\begin{figure}[t]
    \centering
    \includegraphics[width=\linewidth]{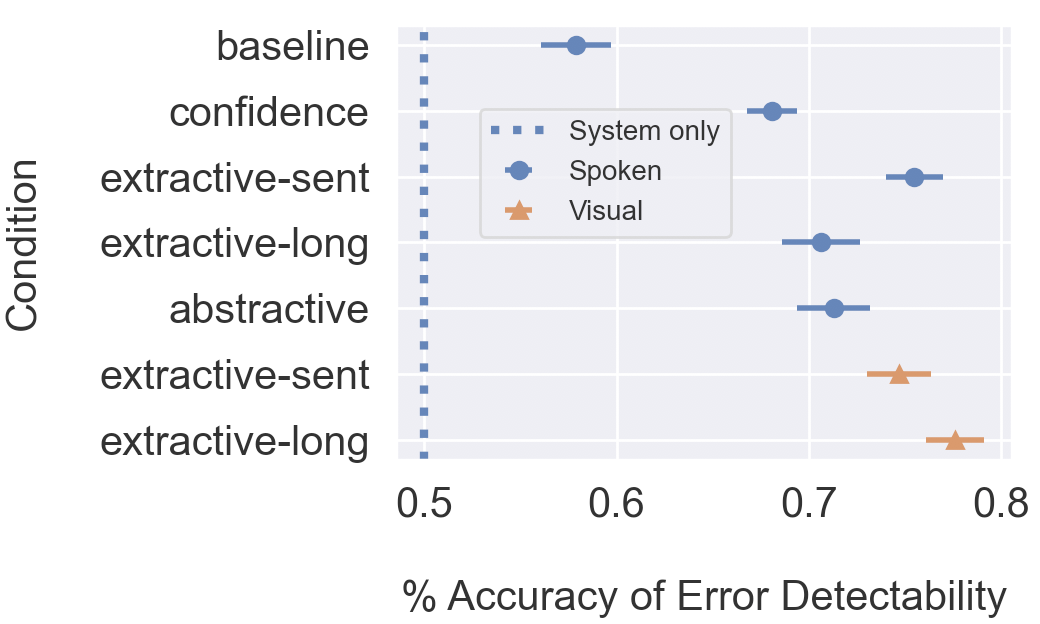}
    \caption{Accuracy of users at \detectError (75 workers per condition). In the  \textit{spoken modality}, {\sc extractive-sent} explanations yield the best results and is significantly better than {\sc confidence}. In contrast, in the \textit{visual modality}, {\sc extractive-long} explanations perform best. We observe a statistically significant ($p<0.01$) difference between {\sc extractive-long} in visual vs spoken, perhaps due to differences in user's cognitive limitations across modalities.}
    \label{fig:result:accuracy}
\end{figure}

\section{Results}

To validate our hypothesis (Section~\ref{sec:hyp}) we compare explanation methods on the quantitative metrics (Section~\ref{sec:quant}). To further understand participant behavior we analyze responses to the post-task survey (Section~\ref{sec:qual}), and analyze common cases where explanations misled the users (Section~\ref{sec:mislead}). Results for reward and time metrics are included in Appendix \ref{sec:reward} and \ref{sec:time}.

\subsection{Quantitative Results}
\label{sec:quant}
Figure~\ref{fig:result:accuracy} shows average user accuracy at \detectError with 75 workers per condition. Similarly to \citet{lamm2020qed}, in order to validate hypotheses and compute statistical significance, we fit a generalized linear mixed effects model using the \verb|lme4| library in R and the formula \verb#a~  c +(1|w) + (1|q)#, where \verb|a| is accuracy, \verb|c| is the condition,  \verb|w| is the worker id and \verb|q| is the question id.   We run pairwise comparisons of these effects using Holm-Bonferroni to correct for multiple hypothesis testing.
For both the spoken and visual modalities, all conditions lead to significantly higher accuracies than {\sc baseline} ($p<0.01$).

\paragraph{Model confidence improved accuracy of \detectError}
In Figure~\ref{fig:result:accuracy}, {\sc confidence} achieves higher accuracy than {\sc baseline}-- 68.1\% vs. 57.2\%. This difference was statistically significant ($p<0.01$), thus validating \textbf{H1}. 
While previous guidelines recommend displaying confidence to users~\cite{amershi-chi19} and show its benefit for sentiment classification and LSAT~\cite{bansal2020does}, our observations provide the first empirical evidence that confidence serves as a simple yet stronger baseline against which explanations for \taskAbr should be compared.

\paragraph{Explaining via an evidence sentence further improved performance.} 
The more interesting comparisons are between explanation types and {\sc confidence}. In both modalities, {\sc extractive-sent} performed better than {\sc confidence}.
 For example, in the \textit{spoken modality}, {\sc extractive-sent} improved accuracy over {\sc confidence}  from 68.1\% to 75.6\% ($p<0.01$); thus validating \textbf{H2}. Contrary to recent prior works that observed no benefit from explaining predictions, this result provides and confirms a concrete application of explanations where they help users in an end-to-end task .

\paragraph{While longer explanations improved performance over more concise explanations in the visual modality, they worsened performance in the spoken modality.}
Figure~\ref{fig:result:accuracy} shows that, for the visual modality, {\sc extractive-long} outperforms {\sc extractive-sent} explanations in the visual modality -- 77.6\% vs. 74.7\% ($p<0.4$). Conversely, for spoken, {\sc extractive-sent} is better than {\sc extractive-long}-- 75.6\% vs. 70.4\% ($p<0.01$); thus supporting \textbf{H3}. In fact, the decrease was severe enough that we no longer observed a statistically significant difference between long explanations and simply communicating confidence ($p=0.9$).

Although communicating the same content, {\em visual} {\sc extractive-long} led to significantly better accuracy than their spoken version---  77.6\% vs. 70.4\% ($p<0.01$); thus validating \textbf{H5}. These results indicate large differences, across modalities, in user ability to process and utilize explanations, and how these differences need to be accounted for while evaluating and developing explanations.

\paragraph{Despite improving conciseness, abstractive summaries (of the longer explanations) did not help improve performance in the spoken modality.}
Figure~\ref{fig:result:accuracy} shows that while {\sc abstractive} performs marginally better than confidence-- 71.3\% vs. 68.1\%, the difference is not statistically significant ($p=0.4$) and thus we could not validate \textbf{H4}. 
This results indicates that the length of the explanation (\eg, number of tokens) is not the only factor that affects user performance, instead the density of information also increases cognitive load on users. This finding is in line with the Time Based Resource Sharing (TBRS) model \citep{barrouillet2007time}, a theory on working memory establishing that time as well as the complexity of what is being communicated, both play a role in cognitive demand. We also observe the same effect in users subjective rating of length of explanation~(Section~\ref{sec:qual}).

\begin{figure}[t]
    \centering
    \includegraphics[width=\linewidth]{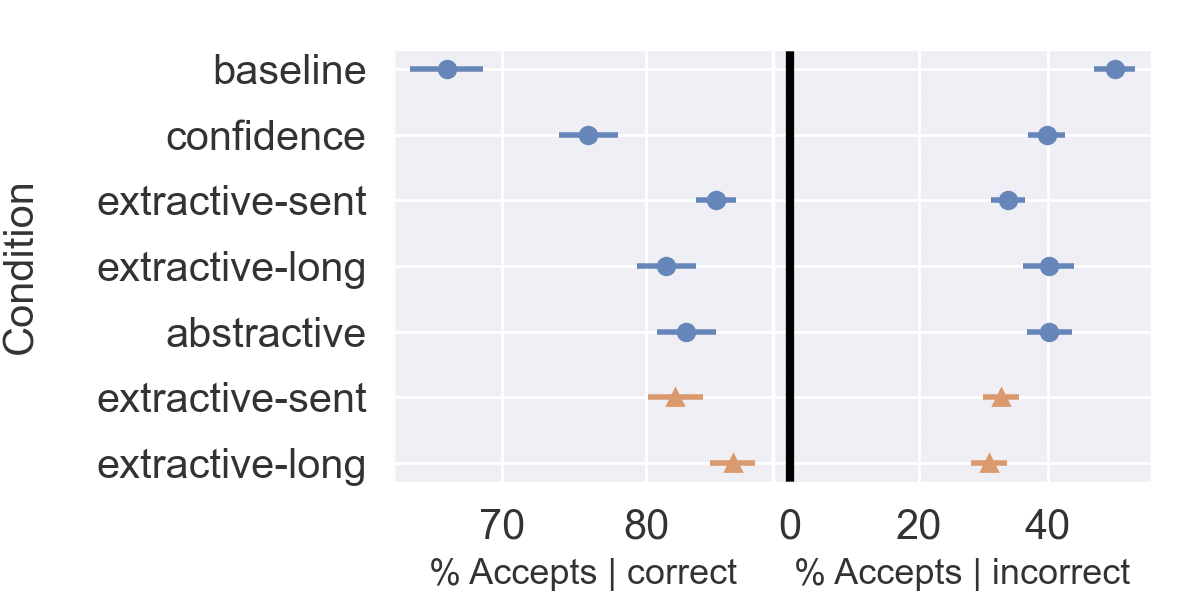}
    \caption{(Left) Explanations significantly increased participant ability to detect \textit{correct} answers compared to simply displaying confidence. (Right) However, only {\sc extractive-sent} in the spoken modality and both explanations in the visual modality decreased the rate at which users are misled.}
    \label{fig:result:fpr}
\end{figure}

\paragraph{All explanations significantly increased participants' ability to detect {\em correct} answers, but only some explanations improved their ability to detect {\em incorrect} answers.}
Instead of aggregate accuracy, Figure~\ref{fig:result:fpr} splits and visualizes how often users accept correct and incorrect answers.
For accepting {\em correct} model predictions, all \textit{visual} and {\em spoken} explanation conditions signficantly helped compared to {\sc confidence} (at least $p<0.05$).

In terms of accepting incorrect predictions, in the \textit{spoken modality}, only {\sc extractive-sent} is significantly better (\ie, lower) than {\sc confidence}---34\% vs. 40\% ($p<0.05$).
Whereas in the \textit{visual modality}, both {\sc extractive-long} and {\sc extractive-sent} lead to improvements over {\sc confidence}--- 30\% ($p<0.01$) and 32\% ($p<0.05$), respectively. This shows that although explanations decrease the chance of being misled by the system, the least misleading explanations change with modality. 

\subsection{Qualitative results}
\label{sec:qual}
We analyzed user responses to the post-task survey to understand their experience, what helped them and how the system could serve them better.

\paragraph{Voice clarity}
To verify that the quality of the text-to-speech tool that we employed did not negatively affect our experiments, we asked users to rate the clarity of the assistant's voice as \textit{very poor, poor, fair, good}, or \textit{excellent}. More than 90\% of participants felt that the voice was good or excellent. 

\paragraph{Length preference}

\begin{figure}[t]
\centering
    \includegraphics[width=.49\textwidth]{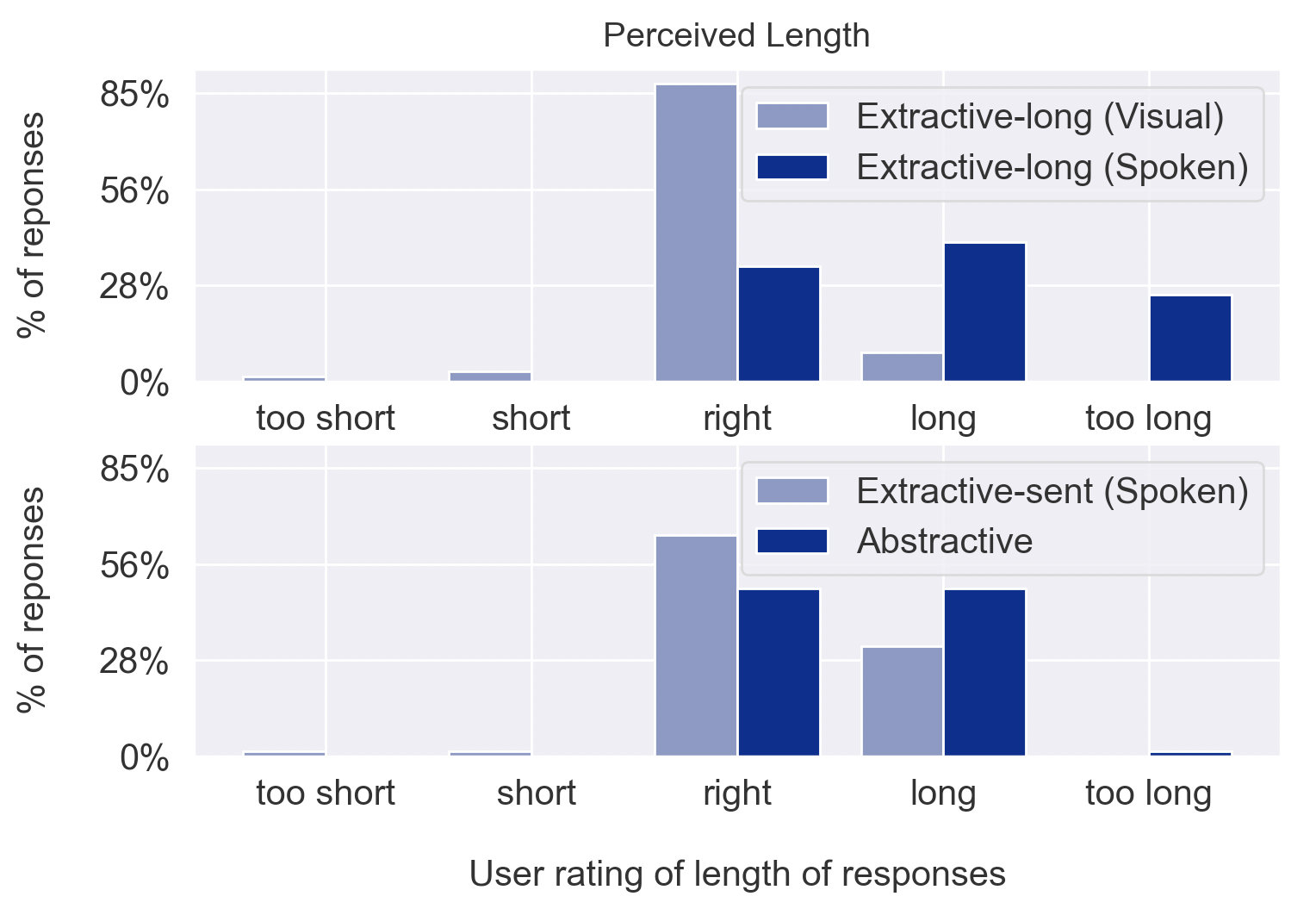}
    \caption{\textbf{Top: } Users perceive the same explanation to be longer in the spoken modality. \textbf{Bottom: }While {\sc extractive-sent} and {\sc abstractive} were the same length, participants rate the latter as longer more often perhaps because of they contain more content.}
    \label{fig:length}
\end{figure}

We asked participants to rate the length of the explanation as \textit{too short, short, right, long,} or \textit{too-long}.
For {\sc extractive-long}, over 85\% of workers perceived that  in the \textit{visual modality}, responses were the right length. On the other hand, in the \textit{spoken modality}, nearly  65\% of participants felt that the responses were too long. In both modalities, users were presented with the same explanations, hence their lengths were exactly the same. This result indicates that what works well in one modality \textit{cannot} simply be transferred as is to another modality.

Additionally, even though {\sc abstractive} and {\sc extractive-sent} were the same duration, in the post experimental survey, users indicated more often that they found {\sc abstractive} to be long, as opposed to {\sc extractive-sent}. As previously mentioned, this relates to the TBRS model of working memory \cite{barrouillet2007time}. We hypothesize that {\em our} {\sc abstractive} explanations, which integrate more information than {\sc extractive-sent} in the same amount of time, might be more taxing in the working memory, making them less effective and in turn making users perceive them as being longer. These results are shown in Figure \ref{fig:length}.

\paragraph{Perceived helpfulness}
Participants were asked whether the responses helped them in their decision making. Their responses showed that {\sc confidence} and all explanation conditions were perceived as helpful by at least 80\% of participants, with no real differences among them except for {\sc extractive-long} in the visual modality (which is perceived helpful by close to 90\% of users). Interestingly, 50\% of participants indicated {\sc baseline} to be helpful. In contrast, our results in Figure~\ref{fig:result:accuracy} show that different explanations actually differ in their eventual helpfulness. These results suggest that subjective measures can sometimes correlate with actual performance when the differences are large, but for the most-part and smaller differences, the result from subjective rating can be unreliable. These findings align with prior observation made \cite{buccinca2020proxy} that showed that evaluating explanations on proxy metrics can lead to incorrect conclusions. More on our these findings are shown in Figure \ref{fig:helpful} in Appendix \ref{sec:helpful}.

\paragraph{User feedback}

In order to get more specific details about how to improve the presentation of information, we asked participants at the end of the survey: \textit{Do you have any additional feedback on what the system can improve?} Two annotators read through about 400 responses across all conditions, and created codes to capture possible areas of improvement. The annotators then used these codes to classify responses. Many users gave feedback that was not insightful (\eg "can't think of anything to improve").  After removing such responses, 175 responses remained for the final analysis. We computed the inter-annotator agreement using Cohen's $k$ ($k$=0.74). Here we briefly describe the most interesting findings, with more details about the codes we used and additional results in Appendix \ref{user-feedback}.

 In {\sc baseline}, where the answer was provided with no additional information, about 50\% of \textbf{participants mentioned that they would have liked it if the voice changed with system certainty}. In {\sc confidence}, around 30\% of participants give this feedback as well. Interestingly, for explanation conditions, this feedback is not seen as often.

For {\sc extractive-sent} in both modalities, {\sc extractive-long} in the visual modality and {\sc abstractive}, 10-35 \% of \textbf{participants would like the level of detail to adapt to the model certainty }. More specifically, users would like to have more details or additional answers \textit{only} when the model is not confident in the prediction. This strategy seems similar to {\em adaptive explanations} proposed by \cite{bansal2020does}.

For the {\sc extractive-long} condition in the \textit{spoken modality} the feedback was mostly about the length of the responses. 78\% of participants mentioned that responses should be shorter, which aligns with the higher perceived length of the explanations in Figure~\ref{fig:length}. For the \textit{visual modality}, 40 \% of  participants mention that highlighting some key items would have made it even easier and faster. In fact, introducing highlights would improve the visual interface, and would likely increase the differences in modalities that we already observe.

Finally, for all explanation conditions, 20-45 \% of \textbf{participants would like to see more than one source containing an answer}. This means that the system would need to find multiple sources that converge to the provided answer. To provide users with this additional information without overloading cognitive capacity, an interactive strategy can be adopted. For example, evidence and additional sources can be presented through an explanatory dialogue \cite{miller2019explanation}, where users are initially provided with limited information, and more can be provided upon request.

\subsection{What misleads users?}
\label{sec:mislead}
To better understand how explanations mislead users and how they can be further improved, we analyzed cases leading to user error. We compiled a set of unique questions alongside their frequency of errors across users in all explanation conditions. 

A single annotator followed a similar coding procedure as previously described, where questions were analyzed in order to detect emergent error categories. Following this initial analysis, questions were categorized into error types.  We found that users tend to be misled on the same questions, with most of the errors happening on around 50 questions per condition, and about 40 of these questions overlapping across conditions. 
 
Below we describe the three main cases:

\paragraph{Plausible explanations.}
A concept is plausible if it is conceptually consistent to what is expected or appropriate in some context \cite{connell2006model}. Work has consistently identified that people often fail to evaluate the accuracy of information  \cite{fazio2008slowing,marsh2013knowledge,fan2020generating}, particularly when no prior knowledge exists and information seems plausible \cite{hinze2014pilgrims}.
We find many cases where a model response and explanation do not answer the question, yet the plausibility misleads users into accepting incorrect responses. For example:

\begin{mdframed}
\small
\textbf{Question: } {\em Who is the patron saint of adoptive parents?}\\

\noindent\textbf{Response:} I am 37 percent confident that the answer is, \textbf{Saint Anthony of Padua}. I found the following 
evidence in a wikipedia passage titled, Anthony of Padua: Saint Anthony of Padua, born Fernando Martins de Bulhoes, also known as Anthony of Lisbon, was a portuguese catholic priest
 and friar of the Franciscan order.

\end{mdframed}

Such errors make up 60 to 65\% of the errors for all explanation conditions.

\paragraph{Lexical overlap.} \citet{mccoy-acl19} describe 3 main heuristics exploited by NLI models: among them, lexical overlap. In our error analysis, the second most common mistake (from 30 to 35\% of errors) that both \textit{the model} and \textit{the users} make is related to the lexical overlap between the question and the evidence. For example:

\begin{mdframed}
\small
\textbf{Question:} {\em How many teams are in the MLB national League?}\\

\noindent\textbf{Response:} I am 60 percent confident that the answer is, \textbf{30}. I found the following evidence in a wikipedia passage titled, Major League Baseball: \textit{A total of 30 teams play in the National League( NL)} and American League (AL) , with 15 teams in each league .
\end{mdframed}

The evidence contains the correct answer (15 teams) but many users are misled by the phrase \textit{``A total of 30 teams play in the National League''}.

\paragraph{Belief bias.}
Humans tend to rely on prior belief when performing reasoning tasks \cite{Klauer:ea:00}. In model evaluation, this has consequences affecting validity. For example, if instructions are not specific, participants are left to use their beliefs to infer what is required of them, leading to varied interpretations. People often rely more heavily on belief bias when processing information under pressure \cite{evans2005rapid}, therefore in time limited evaluations this phenomenon might be more prominent. We reduced these confounds by carefully designing instructions, a straightforward interface, allowing workers plenty of time and removing ambiguous questions. 
However, some interesting cases of belief bias do occur --- take for instance, the example below:
    
\begin{mdframed}
\small
\textbf{Question: } {\em \textbf{Where} is the longest bone in the body found?} \\

\noindent \textbf{Response: } I am 17 percent confident that the answer is, \textbf{femur}. I found the following evidence in a wikipedia passage titled, Femur:
The Femur or thigh bone, is the proximal bone of the hindlimb in tetrapod vertebrates .

\end{mdframed}

Such errors in our evaluation make about 3-5\% of total errors in each explanation condition.

\section{Discussion}

\subsection{Why Explanations Worked for \taskAbr?}
Our studies observed significant improvements from explanations for the end-task to help users decide whether to trust the prediction of an imperfect open-domain QA agent.
A natural question then is why explanations worked for this task despite many negative results on other tasks, such as sentiment, LSAT, and computer vision~\cite{bansal2020does,chu2020visual,hase2020evaluating}.
One hypothesis is that, in our task
explanations provide user's with new knowledge, previously unknown to them.
For example, on a sentiment classification task, explanations highlight information in the text already visible to the users.
In contrast, in \taskAbr, explanations in terms of evidence text provide users with the additional context that helps validate the system's answers. This new evidence may be especially helpful to users who don't know the answers to the question, which would often be the case for open-domain question answering.

However, its worth noting that like previous works, not all of our explanation methods provided significant value over confidence; \eg, in Figure~\ref{fig:result:accuracy} we did not observe any significant differences between longer extractive explanations and confidence for the spoken modality. Thus the success from showing explanations still cannot be taken for granted but instead be measured using well-designed user studies.

\subsection{Implications and Recommendations}

Another interesting question is how can our findings inform future research in explainable NLP. 

\paragraph{Develop modality-specific explanations}
Our results showed that the best explanation varied across the modalities, indicating that evaluating explanations on one modality (\eg, using studies with visual UI) and deploying them on another (\eg, voice assistant) can lead to sub-optimal deployment decisions. As a result, explanations should be evaluated in the task and settings in which they will be deployed in-the-wild.

\paragraph{Further study abstractive explanations}
Longer explanations helped in the visual case, showing that communicating more evidence has potential to help users. But, they hurt in the spoken case, perhaps because longer explanations increase cognitive load on users.
This may indicate a trade-off between {\em information content} of explanation and its {\em cognitive load} for \taskAbr.
We had hoped abstractive explanations would achieve a more optimal balance between fidelity and comprehensibility for spoken.
However, Figure~\ref{fig:result:accuracy} shows that they did not improve end-performance. One reason is that even though abstractive explanations were concise, they had high information density and thus did not sufficiently decrease cognitive load.

That said, while abstractive explanations did not significantly improve accuracy compared to longer explanations, they did  improve user speed at the task by 2.2 sec (Table~\ref{tab:time}) and were satisfactory rated in terms of their perceived length compared to longer explanations~(Figure~\ref{fig:length}). The utility of such generated explanations, over longer explanations, may further increase when explain multiple sources (\eg, in Hotpot QA \cite{yang2018hotpotqa}) or candidate answers, where communicating multiple entire passages seems infeasible.

\paragraph{Enable interactive explanations}
All explanation conditions we tested were static-- they assumed a single trade-off between detail and conciseness. For example, {\sc extractive-sent} always conveyed a single sentence to the user as an explanation, which was concise but may not always convey all context required to validate answers.
A different strategy may be to use interactive explanations, 
in which the system first gives a concise explanation and then lets users request additional information. 
Such explanations may be especially used to accommodate user suggestions such as, including and explaining multiple candidate answers or multiple answers sources. 
Another possible strategy is to use \textit{adaptive explanations}, where the model switches explanation strategies based on its confidence~\cite{bansal2020does}.

\paragraph{Limitations}

While our user study addresses issues of many similar previous evaluations of explanations, it still has limitations. First, although the interaction of users with the QA system was kept as realistic as possible, in reality users may have the option to double-check the model's answer using external tools, such as Web search. Accommodating that case, would require encoding the additional cost of the reject action (\eg, due to time spent and effort) into the payoff. In addition, unlike an interaction in-the-wild, questions were not posed by participants themselves, which may lead to different kinds of biases in the interpretation of the questions, as discussed before. Second, we conducted studies with MTurk workers who may not have the same motivation for performing the task as real users. To address this we incentivized them by rewarding high-performance through bonuses and penalties specified using a payoff matrix. In practice, the values of the payoff matrix can vary depending on the stake of the domain and may vary with users. Finally, we only registered hypothesis that compared performance on one metric-- accuracy of \detectError. However, there may be other metrics that may be of interest, \eg, improvements in speed and user satisfaction.

\section{Conclusion}

Contrary to recent user-studies for other tasks (such as classification), ours suggest that for \taskAbr, explanations of model's predictions significantly help end-users decide when to trust the model's answers, over strong baselines such as displaying calibrated confidence.
We observed this for two scenarios where users interact with \taskAbr model using spoken or visual modalities.
However, the best explanation may change with the modality, \eg, due to differences in users' cognitive abilities across modalities.
For example, for the spoken modality, concise explanations that highlight the sentence containing the answer worked well. In contrast, for the visual modality, performance improved upon showing longer explanations.
Thus, developers and researchers of explainable \taskAbr systems should evaluate explanations on the task and modalities where these models will be eventually deployed, and tune these explanations while accounting for user needs and limitations.

Despite success of explanations on our domain, explanations sometimes still mislead users into trusting an incorrect prediction, and sometimes as often as displaying the baselines. These results indicate the need to develop better explanations or other mechanisms to further appropriate user reliance on \taskAbr agents, \eg, by enabling abstractive explanations that balance conciseness and detail while taking into account user's cognitive limitations, interactive explanations that can explain multiple answer sources and candidates and adaptive explanations where model strategy changes based on its confidence.

\bibliography{anthology,custom}
\bibliographystyle{acl_natbib}

\clearpage
\appendix
\section{Explanation Examples}
\label{sec:appendix1}
In Table \ref{tab:examples}, we show an example of how the responses and explanations looked for each of the conditions. We also indicate in which modalities each explanation is shown in our experiments.

\begin{table*}[t]
\begin{tabular}{@{}lp{10cm}l@{}}
\toprule
{\sc Explanation Type} &
  {\sc Response+Explanation} &
  {\sc Modality} \\ \midrule
  
Baseline &
  The answer is, \textbf{two}. &
  Spoken \\
  \midrule
Confidence &
  I am 41 percent confident that the answer is, \textbf{ two}. &
  Spoken \\
  \midrule
Abstractive &
  I am 41 percent confident that the answer is, \textbf{two}. I summarized evidence from  a wikipedia passage titled, Marco Polo (TV series). Netflix cancelled the show after two seasons, as it had resulted in a  200 million dollar loss. &
  Spoken \\
  \midrule
Extractive-sent &
  I am 41 percent confident that the answer is, \textbf{two}. I found the following evidence in a wikipedia passage titled, Marco Polo (TV series). On December 12, 2016, Netflix announced they had canceled "Marco Polo" after two seasons. &
  Spoken/visual. \\
  \midrule
Extractive-long &
  I am 41 percent confident that the answer is, \textbf{two}. I found the following evidence in a wikipedia passage titled, Marco Polo (TV series).  On December 12, 2016, Netflix announced they had canceled "Marco Polo " after two seasons. Sources told "The Hollywood Reporter" that the series' two seasons resulted in a  200 million dollar loss for Netflix , and the decision to cancel the series was jointly taken by Netflix and the Weinstein Company. Luthi portrays Ling Ling in season 1, Chew in season 2. The series was originally developed at starz, which had picked up the series in January 2012. &
  Spoken/visual \\ \bottomrule
\end{tabular}
\caption{\textbf{Explanation examples:} Example of how system responses looked for each explanation type and baseline, for the question \textit{How many seasons of Marco Polo are there?} }
\label{tab:examples}
\end{table*}

\section{Temperature Scaling}
\label{sec:temp-scale}

Temperature scaling \cite{guo2017calibration}, a multiclass extension of Platt Scaling \cite{platt1999probabilistic}, is a post-processing method applied on the logits of a neural network, before the softmax layer. It consists of learning a scalar parameter $t$, which decreases or increases confidence. $t$ is used to rescale the logit vector $z$, which is input to softmax $\sigma$, so that the predicted probabilities are obtained by $\sigma(z/t)$, instead of $\sigma(z)$.

In our experiments, the model is set to pick from the top 100 solutions, however, in many cases the correct answer occurs within the top 10 items. For our purposes we calibrate the confidence scores of the top 10 outputs. We use the publicly available scripts provided by \citet{guo2017calibration}.\footnote{\url{https://github.com/gpleiss/temperature_scaling}}

\begin{figure}[h]
    \centering
    \includegraphics[width=\linewidth]{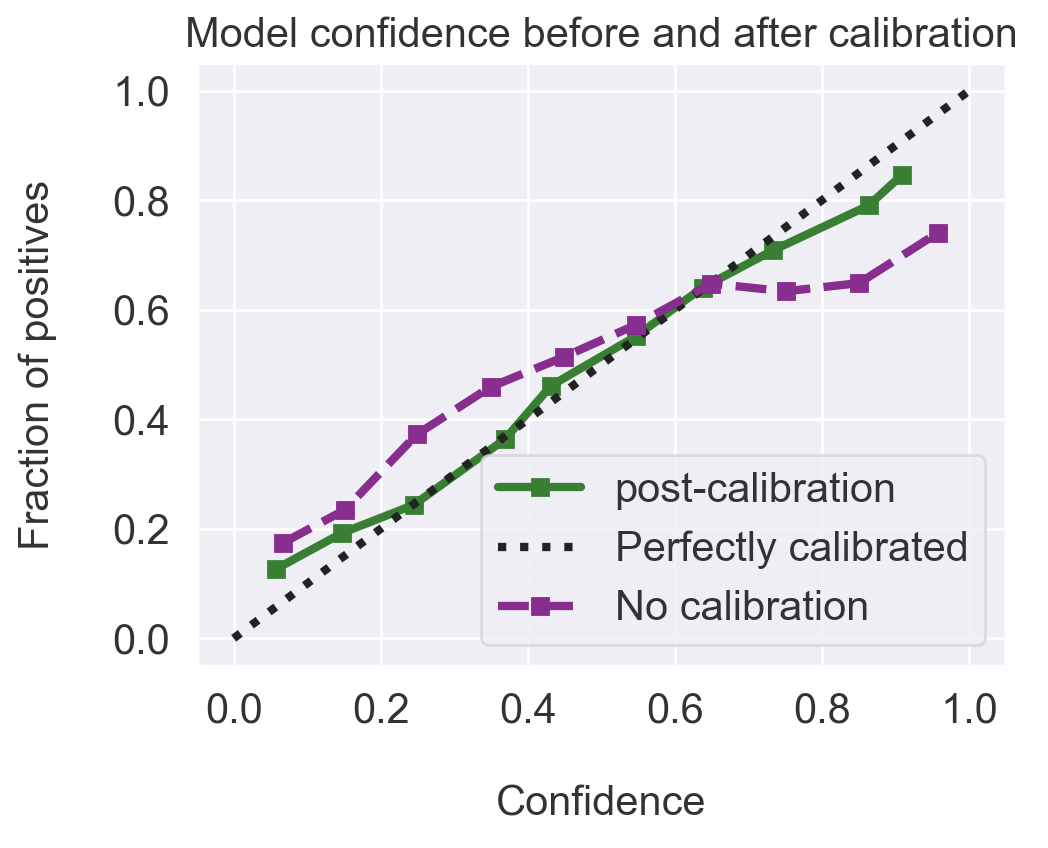}
    \caption{Confidence before and after calibration.}
    \label{fig:calibration}
    
\end{figure}

The model confidence before and after calibration can be seen in Figure \ref{fig:calibration}.

\section{Additional Preprocessing}
\label{sec:preprocess}

Additional preprocessing to ascertain the quality of stimuli in each modality was required; more details can be found in Appendix .
Before sampling questions for the task, to ensure a high-quality and non-ambiguous experience for MTurk workers, we manually filter out several ``problematic'' questions:


\begin{itemize}
    \item \textbf{Ambiguity in the question}: For various questions in NQ, multiple answers can exist. For example, the question:  \textit{when was King Kong released?}, does not specify which of the many King Kong movies or video games it refers to. These cases have been known to appear often in NQ \cite{min2020ambigqa}. We remove such questions from our subset.
    
    \item \textbf{The gold answer was incorrect}: Many examples in NQ are incorrectly annotated. As it is too expensive to re annotate these cases, we remove them.
    
    \item \textbf{Answer marked incorrect is actually correct} : We present both correct and incorrect questions to users. There are cases where the predicted answer is marked incorrect (not exact match) but is actually correct (a paraphrase). We manually verify that correct answers are paired with contexts which support the answer.
    
    \item \textbf{Correct answer but incorrect evidence}: The model sometimes, though not as often, chooses the correct answer but in the incorrect context. We discarded examples where the explanation was irrelevant to the question e.g. who plays Oscar in the office? \textit{Oscar Nuñez, is a Cuban-American actor and comedian.}. In order to be able to make more general conclusions about whether explanations help in \detectError, we restrict our questions to ones containing correct answers in the correct context.
    
    \item \textbf{Question and prediction do not match type}. We removed cases where the question asked for a certain type e.g. a date, and the prediction type did not match e.g. a location.   
    
\end{itemize}

In the visual modality, to ensure readability, we fixed capitalizations. For the spoken modality, to ensure fluency and clarity, we manually (1) inserted punctuation to ensure more natural sounding pauses, and (2) changed abbreviations and symbols to a written out form e.g. \textit{\$ 3.5 billion } to \textit{3.5 billion dollars}.

\section{Task Setup: Additional details}
\label{sec:appendix2}
\paragraph{Platform and participant details}

We conduct our experiments using Amazon Mechanical Turk\footnote{\url{https://www.mturk.com/}}.  We recruited 525 participants in total, with approval ratings greater than 95 \% and had a maximum of 8 days for approval of responses in order to minimize the amount of spamming.

We use a random sample of 120 questions from our dataset which remains the same across all conditions. In order to keep each session per participant at a reasonable time and ensure the quality of the data wouldn't be affected by workers becoming exhausted,  we opted for three fixed batches of 40 questions, all split as 50 \% correct and 50 \% incorrect. Workers could only participate once (only one batch in one condition). Participants took around from 35-45 minutes to complete the HITs, but were given up to 70 minutes to complete.

We monitored if their screen went out of focus, to ensure that participants did not cheat. We ensured that we had 25 user annotations per question. When analyzing the data, we remove the first 4 questions of each batch, as it may take participants a few tries before getting used to the interface. In the end, we collect about 21,000 test instances.

\paragraph{Task Instructions}
Imagine asking Norby a question and Norby responds with an answer. Norby's answer can be correct or wrong. If you believe Norby's answer is correct, you can accept the answer. If you believe it is wrong, you can reject it. If the answer is actually correct and you accept it, you will earn a bonus of \$0.15. But, if the answer is wrong, and you accept it, you will lose \$0.15 from your bonus. If you reject the answer, your bonus is not affected. (Don't worry, the bonus is extra! Even if it shows negative during the experiment, in the end the minimum bonus is 0). In total you will see 40 questions in this HIT (you will only be allowed to participate once) and the task will take about 40 to 45 minutes. You can be compensated a maximum of \$13.50 for about 40-45 minutes of work.
Some things to note:
\begin{enumerate}
    \item You must listen to the audio before the options become available.
    
    \item If you make it to the end there is a submit button there, however, in case of an emergency you can hit the quit early button above and you will get rewarded for the answers you provided.
    
    \item You can play the audio as many times as you need but as soon as you click a choice you will be directed to the next item.
    
    \item \textbf{IMPORTANT!!} Please do not look up questions in any search engine. We will monitor when the screen goes out of focus, so please keep the screen on focus or you might risk being rejected.

    \item Finally, please do not discuss answers in forums; that will invalidate our results.
\end{enumerate}

\section{Post-task survey}
\label{sec:survey}

\begin{enumerate}
    \item I found the CLARITY of Norby's voice to be:
    
    \textbf{(a)} Excellent \textbf{(b)} Good \textbf{(c)} Fair \textbf{(d)} Poor \textbf{(e)} Very Poor
    
    \item I found Norby's responses to be HELPFUL when deciding to Accept or Reject:
    
    \textbf{(a)} Strongly Agree \textbf{(b)} Agree \textbf{(c)} Undecided \textbf{(d)} Disagree \textbf{(e)} Strongly Disagree
    
    Can you give a few more details about your answer?
    
    \item I found the LENGTH of Norby's responses to be:
    
    \textbf{(a)} Too Long  \textbf{(b)} Long  \textbf{(c)} Just right \textbf{(d)} Short \textbf{(e)} Too short
    
    \item No AI is perfect and Norby is no exception. We are interested in helping Norby provide responses that can help users to determine whether to trust it or not (to accept or reject, just as you have done in this experiment). From your interaction with Norby, \textbf{do you have any additional feedback on what it can improve?}
\end{enumerate}

\section{Results}

\subsection{Reward}
\begin{figure}[h]
\centering
    \includegraphics[width=.48\textwidth]{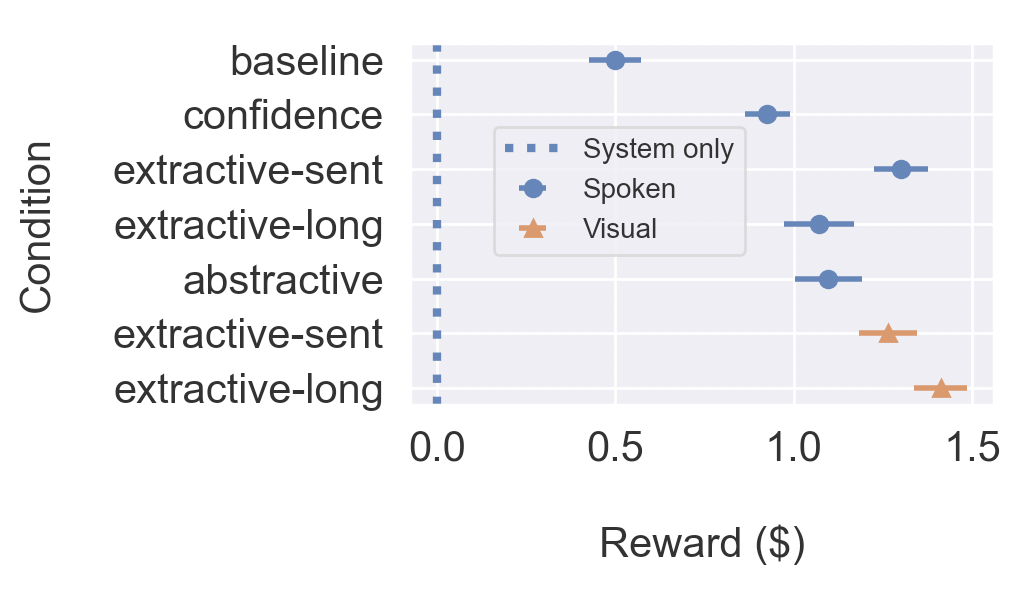}
    \caption{\textbf{Reward}: The scores presented here are out of \$ 2.70. Although all explanations are better than {\sc confidence}, the explanations leading to the highest rewards change across modalities. }
    \label{fig:reward}
\end{figure}

We compute the differences in overall reward for each condition. We observe the same trends as we discussed for accuracy. More specifically, all explanation conditions improve the final user reward, with {\sc extractive-sent} performing best in the spoken modality and {\sc extractive-long} performing best overall. These differences are shown in Figure \ref{fig:reward}.

\label{sec:reward}

\subsection{Time Differences}
\label{sec:time}
We measured the time (in seconds) that it took participants to complete each question. In Table \ref{tab:time} we present the median times averaged over all workers per condition. We also include an adjusted time, subtracting the length of the audio, in order to measure decision time. 
\begin{table}[h]

\centering
\begin{tabular}{lll}
\toprule
{\sc condition} & {\sc sec/question} & {\sc adjusted} \\ \midrule
\multicolumn{3}{l}{\sc Spoken Modality} \\ \midrule
Baseline              & 10.2 $\pm$ 1.6       &    8.3 $\pm$ 1.6   \\
Confidence            & 9.4 $\pm$ 1.5       &  6.0 $\pm$ 1.5     \\
Abstractive           & 24.4 $\pm$ 1.5       &  7.0 $\pm$ 1.4     \\
Extractive-long       & 44.9 $\pm$1.6       &   9.2 $\pm$ 1.6    \\
Extractive-sent       & 24.3 $\pm$1.7       &   7.6 $\pm$   1.7  \\ \midrule
\multicolumn{3}{l}{\sc Visual Modality} \\ \midrule
Extractive-long       & 16.1 $\pm$ 1.7       &   -    \\
Extractive-sent       & 10.4$\pm$1.1       &    -  \\ \bottomrule
\end{tabular}
\caption{Time differences across modalities. Time differences in the right column have been adjusted by removing the duration of the audio files. We observe that with additional information, users can make faster decisions than the {\sc baseline } condition. }
\label{tab:time}
\end{table}

\subsection{Voice Quality}
\label{sec:voice}
\begin{figure}[h]
\centering
    \includegraphics[width=.5\textwidth]{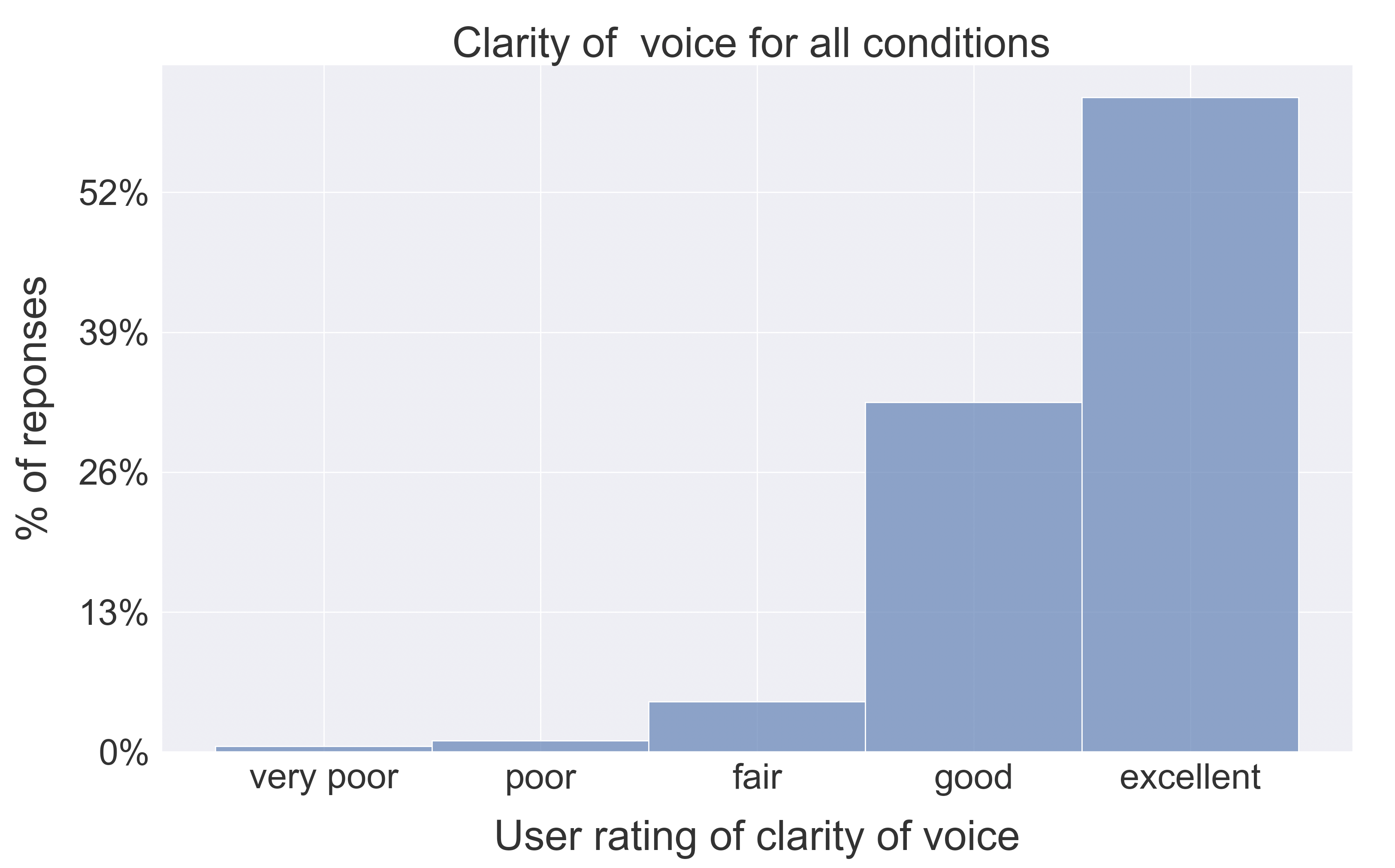}
    \caption{\textbf{Voice clarity: } Most participants found the voice of the assistant to be good or excellent.}
    \label{fig:voice}
\end{figure}

Participants in the spoken conditions rated how clear they found the voice to be. Around 90 \% rated the voice as good or excellent. These results are shown in Figure \ref{fig:voice}.

\subsection{Helpfulness}
\label{sec:helpful}

\begin{figure}[h!]
\centering
    \includegraphics[width=.5\textwidth]{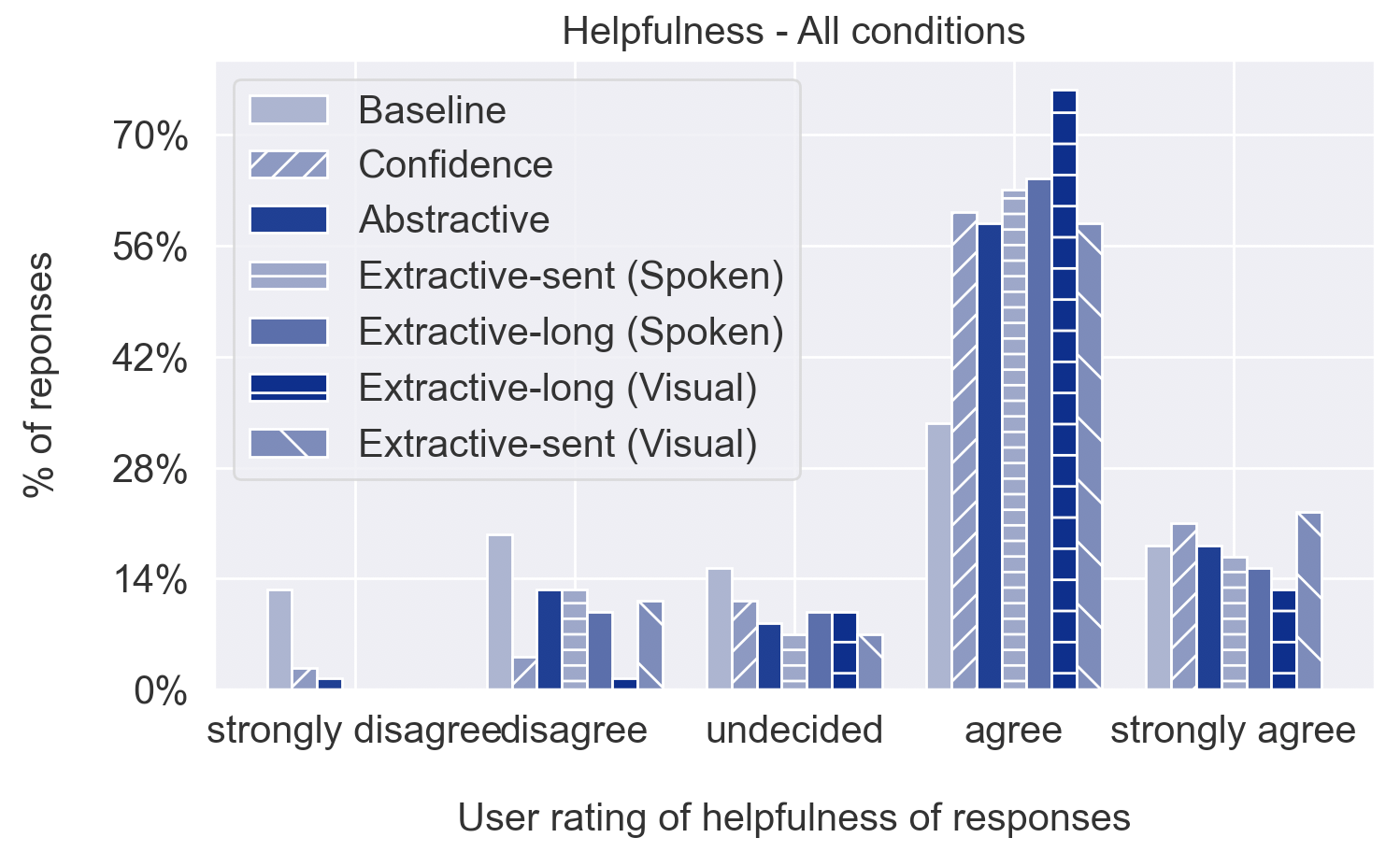}
    \caption{ \textbf{Helpfulness:} Participants indicated how helpful responses were. These results reflect the large differences we see in performance ({\sc baseline } vs the rest of the settings), but are not able to capture the more subtle differences among explanation strategies and {\sc confidence}.}
    \label{fig:helpful}
\end{figure}

Differences in perceived helpfulness are shown in Figure \ref{fig:helpful}.

\subsection{User Feedback}
\begin{table*}[h]
\centering
\begin{tabular}{lp{6cm}p{4cm}}
\toprule
{\sc Code}     & {\sc Description}                                         & {\sc Category}                            \\
\midrule
len-conciseness   & users wish explanation was shorter              & \multirow{2}{*}{improvement on length}\\
len-expand        & users wish explanation was shorter         &  \\
\midrule
adapt-detail &
  users wish details adapted with confidence  &
  \multirow{2}{*}{adaptability feature } \\
adapt-voice       & users wish voice adapted to confidence &                         \\
\midrule
pres-change-confidence &
  users wish confidence would be communicated differently e.g. the answer is probably.... &
  \multirow{2}{*}{improve presentation} \\
pres-highlighting &
  users wish important facts would be highlighted &
   \\
\midrule
need-more-sources & users wish more source were provided            & \\
need-confidence   & users wish confidence was provided           &  \\
need-source       & users wished a source was provided                           &  \multirow{5}{*}{need additional info} \\
need-explanation  & users wish an explanation would be provided                &  \\
need-link         & users wish a link was provided                     &  \\
need-multiple-answers &
  users wish more than 1 answer was provided &
 \\
  \bottomrule
\end{tabular}
\caption{The codes used to uncover areas of improvement from the post-experimental user feedback.}
\label{tab:codes-feedback}
\end{table*}

\label{user-feedback}
\begin{table*}[t]
\centering
\begin{tabular}{@{}clc@{}}
\toprule
{\sc condition}                                & {\sc code}                   & \% {\sc participants} \\ \midrule
\multirow{4}{*}{baseline}                 & adapt-voice            & 50              \\
                                          & need-confidence        & 36              \\
                                          & need-explanation       & 25              \\
                                          & need-source            & 17              \\ \midrule
\multirow{5}{*}{confidence}               & need-explanation       & 38              \\
                                          & adapt-voice            & 29              \\
                                          & pres-change-confidence & 14              \\
                                          & adapt-detail           & 10              \\
                                          & need-multiple-answers  & 10              \\
\multicolumn{1}{l}{}                      & need-link              & 5               \\ \midrule
\multirow{7}{*}{extractive-sent (spoken)} & need-more-sources      & 44              \\
                                          & adapt-detail           & 28              \\
                                          & len-conciseness        & 22              \\
                                          & need-multiple-answers  & 17              \\
                                          & need-link              & 11              \\
                                          & len-expand             & 11              \\
                                          & pres-change-confidence & 6               \\ \midrule
\multirow{3}{*}{extractive-long (spoken)} & len-conciseness        & 78              \\
                                          & need-more-sources      & 15              \\
                                          & pres-change-confidence & 4               \\ \midrule
\multirow{5}{*}{abstractive}              & len-conciseness        & 52              \\
                                          & need-more-sources      & 22              \\
                                          & adapt-detail           & 22              \\
                                          & pres-change-confidence & 13              \\
                                          & need-multiple-answers  & 4               \\ \midrule
\multirow{4}{*}{extractive-sent (visual)} & need-more-sources      & 33              \\
                                          & adapt-detail           & 33              \\
                                          & len-expand             & 27              \\
                                          & need-multiple-answers  & 7               \\ \midrule
extractive-long (visual)                  & pres-highlighting      & 40              \\
\multicolumn{1}{l}{}                      & need-more-sources      & 33              \\
\multicolumn{1}{l}{}                      & adapt-detail           & 10              \\
\multicolumn{1}{l}{}                      & need-link              & 10              \\
\multicolumn{1}{l}{}                      & pres-change-confidence & 7   \\
\bottomrule
\end{tabular}
\caption{Distribution of codes across all conditions. Codes are \textbf{not} mutually exclusive.}
\label{tab:coded-responses}
\end{table*}

Users provided free-form written feedback on possible ways to improve the system. The prompt they saw was: \textit{do you have any additional feedback on
what the system can improve?} After converging on a final set of codes, two annotators coded up about 400 responses across all conditions. The codes and their descriptions can be found in Table \ref{tab:codes-feedback}. The codes are \textit{not} mutually exclusive.

 We found that many users across most conditions, would like \textbf{adaptability features} added. Additionally, we found that participants would like to be provided with multiple sources which converge on the answer. We also observe that for spoken conditions, \textbf{improvements on length} are mentioned more often. The full distribution of codes across conditions is shown in Table \ref{tab:coded-responses}.

\end{document}